\documentclass[runningheads]{llncs}
\usepackage{amssymb}
\usepackage{graphicx}
\usepackage{amsmath}
\usepackage{commath}
\usepackage{float}
\usepackage{blindtext}
\usepackage{scrextend}
\usepackage{multirow}
\begin{document}
\title{Do Public Datasets Assure Unbiased Comparisons for Registration Evaluation?}
%

%

\author{Jie Luo\inst{1,2}
	\and Guangshen Ma\inst{3}
	\and Sarah Frisken\inst{1}
	\and Parikshit Juvekar\inst{1} 
	\and Nazim Haouchine\inst{1} 
	\and Zhe Xu\inst{1} 
	\and Yiming Xiao\inst{4}
	\and Alexandra Golby\inst{1} 
	\and Patrick Codd\inst{3} 
	\and Masashi Sugiyama\inst{5,2} 
	\and William Wells III\inst{1} 
}

\authorrunning{J. Luo et al.}

%


\institute{
	Brigham and Women's Hospital, Harvard Medical School, USA 
	\and Graduate School of Frontier Sciences, The University of Tokyo, Japan 
	\and Department of Mechanical Engineering and Material Science, Duke University, USA 
	\and Robarts Research Institute, Western University, Canada
	\and Center for Advanced Intelligence Project, RIKEN, Japan\\
	\email{jluo5@bwh.harvard.edu}
}

\maketitle

\vspace{-6mm}
\begin{abstract}
With the increasing availability of new image registration approaches, an unbiased evaluation is becoming more needed so that clinicians can choose the most suitable approaches for their applications. Current evaluations typically use landmarks in manually annotated datasets. As a result, the quality of annotations is crucial for unbiased comparisons. Even though most data providers claim to have quality control over their datasets, an \textit{objective third-party screening} can be reassuring for intended users. In this study, we use the variogram to screen the manually annotated landmarks in two datasets used to benchmark registration in image-guided neurosurgeries. The variogram provides an intuitive 2D representation of the spatial characteristics of annotated landmarks. Using variograms, we identified potentially problematic cases and had them examined by experienced radiologists. We found that (1) a small number of annotations may have fiducial localization errors; (2) the landmark distribution for some cases is not ideal to offer fair comparisons. If unresolved, both findings could incur bias in registration evaluation.

\end{abstract}
\vspace{-8mm}
\section{Introduction}
\vspace{-1mm}

The evaluation of non-rigid registration is challenging for two reasons: firstly, quantitative validation between aligned images is sometimes difficult due to lack of ground truth. Secondly,  the location where accurate registration is needed may vary by surgical procedure, e.g., in brain atlas building, good alignment of the ventricle region is sought after, while in image-guided neurosurgery, surgeons are interested in accurate registration of the preoperative (\textit{p}-) tumor boundary to the intraoperative (\textit{i}-) coordinate space \cite{Gerard,Morin,Jax,Ma}. Because of these issues, it is difficult to set up a unified standard to characterize registration error \cite{Maintz,Sotiras}. 

\vspace{-3mm}
\subsubsection{Related work} In early work, image similarity measures, such as the sum of squared differences or mutual information, were used as evaluation criteria for registration \cite{Maintz,Sotiras,Song}. The ``Retrospective Image Registration Evaluation" (RIRE) project \cite{Fitzpatrick,Fitzpatrick2,Fitzpatrick3} introduced Target Registration Error (TRE) and Fiducial Registration Error (FRE) to evaluate registration. TRE is the true error of registered target points in physical space, while FRE represents the error of registered annotated fiducial markers in image space. Even though annotated fiducial markers may not be truly accurate due to operator error, FRE is often used as a surrogate of TRE for convenience \cite{Fitzpatrick4,Datteri,Min}. The Vista \cite{Barillot} and NIREP \cite{NIREP} projects included additional registration error measures, e.g., the region of interest (ROI) overlap, the average volume difference and the Jacobian of the deformation field. Recently, multiple registration approaches were compared based on Computed Tomography of the abdomen \cite{Xu,Kabus,Murphy} and Magnetic Resonance (MR) images of the brain \cite{Yassa,Klein1,Klein2,Ou}.

Due to its simplicity and the reliability issues of other criteria \cite{Rohlfing}, FRE has become the most widely used registration error measure. However, using FRE has certain limitations:

\begin{enumerate}
	\item Because fiducial markers (landmarks) are annotated by localization algorithms (manual, automatic or semi-automatic methods), they may contain Fiducial Localization Error (FLE) \cite{Fitzpatrick2}, which measures the discrepancy between an annotated landmark and its true location. FLEs cause false registration errors and should be avoided.	
	\item The FRE only estimates the error at specific landmark locations, thus a dense population of landmarks is preferred. If landmarks are sparse or are not distributed evenly across the entire ROI, a bias that favors regions with landmarks in the registration evaluation may be introduced.
\end{enumerate}

Recently, more annotated data sets are becoming publicly available and these sets are being used to compare existing algorithms and evaluate new methods. Newly proposed registration algorithms are then characterized only by demonstrating superior performance on these datasets \cite{Ines,CURIOUS}.  

To provide an unbiased evaluation of registration, the quality of the annotations is crucial. However, to the best of our knowledge, objective quality control over annotation in public datasets has been overlooked by the image registration community. Even though many providers claim to have mitigated FLEs and other problems by having multiple observers localize the landmarks (and averaging their results) \cite{Song}, an \textit{objective third-party screening} can be reassuring for intended users.

In this study, we perform a third-party screening of the annotations of two benchmark datasets for image-guided neurosurgery, RESECT \cite{RESECT} and BITE \cite{BITE}. Both datasets have corresponding landmarks on \textit{p}-MR and \textit{i}-Ultrasound (US) images. Minimizing the FLE for these landmarks is the standard evaluation method in related registration challenges \cite{CURIOUS}. The tool we choose for the screening is called the variogram, which has been extensively used to describe the spatial dependence of minerals in geostatistics. The variogram has been brought to the medical imaging field as a means to identify vector outliers for landmark-based image registration \cite{JaxVario}. In this screening, we want to (1) detect any obvious FLEs; and (2) examine the distribution of annotated landmarks. We also provide constructive discussion about the impact of our findings.

\vspace{-2mm}
\section{Method}
\vspace{-1mm}

For each pair of annotated images, we compute displacements between pairs of corresponding landmarks to generate a 3D vector field $\mathcal{D}$. By analyzing $\mathcal{D}$, we can assess the quality of the annotations. The variogram characterizes the spatial dependence of $\mathcal{D}$ and provides an intuitive 2D representation for visual inspection \cite{JaxVario}. 

In this section, we review constructing the variogram for a vector field and explain how to use the variogram to flag potential FLEs and problematic landmark distributions.

\vspace{-1mm}
\subsection{Constructing the variogram}

In an image registration example, let $\Omega\subset\mathbb{R}^3$, $I_\mathrm{f}: \Omega\rightarrow\mathbb{R}$ and $I_\mathrm{m}: \Omega\rightarrow\mathbb{R}$ be the fixed and moving images. $(\mathbf{x},\mathbf{x}')\in\Omega$ represents a pair of manually labeled corresponding landmarks on $I_\mathrm{f}$ and $I_\mathrm{m}$. $\mathbf{d}(\mathbf{x})=\mathbf{x}'-\mathbf{x}$ is the displacement vector from $\mathbf{x}$ to $\mathbf{x}'$. For $K$ pairs of landmarks, we have a set of displacement vectors $\mathcal{D} = \{\mathbf{d}(\mathbf{x}_\mathit{k})\}_{\mathit{k}=1}^K$.

Given a landmark location $\mathbf{s}$, let $\mathit{h}$ represent the distance to $\mathbf{s}$. The theoretical variogram  $\gamma(\mathit{h})$ is the expected value of the differences between $\mathbf{d}(\mathbf{s})$ and other $\mathbf{d}$'s whose starting points are $\mathit{h}$ away from $\mathbf{s}$:
\begin{equation}
\gamma(\mathit{h})=\frac{1}{2}E[(\mathbf{d}(\mathbf{s})-\mathbf{d}(\mathbf{s}+\mathit{h}))^2],
\end{equation}
here $\gamma(\mathit{h})$ describes the spatial dependency of displacement vectors as a function of the distance.

In this study, we are interested in the pairwise spatial dependence of all  displacement vectors. Therefore we use the empirical variogram cloud $\hat{\gamma}(\mathit{h})$ instead of $\gamma(\mathit{h})$. Given a vector field $\mathcal{D}$, $\hat{\gamma}(h)$ can be constructed as follows:
\begin{enumerate}
	\item For each pair of vectors  $(\mathbf{d}(\mathbf{x}_\mathit{i}),\mathbf{d}(\mathbf{x}_\mathit{j}))$, compute $\epsilon_\mathit{ij}=\norm{\mathbf{d}(\mathbf{x}_i)-\mathbf{d}(\mathbf{x}_j)}^2$;
	\item Compute $\mathit{h}_\mathrm{ij}=\norm{\mathbf{x}_i-\mathbf{x}_j}$;
	\item Plot all $(\mathit{h}_\mathit{ij},\epsilon_\mathit{ij})$ to obtain $\hat{\gamma}(h)$.
\end{enumerate}

\begin{figure}[t]
		\vspace{-5mm}
	\centering
	\includegraphics[height=2.7cm]{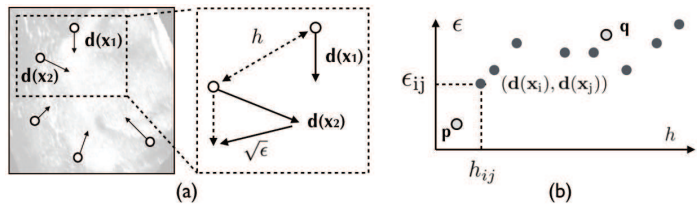}
	\vspace{-3mm}
	\caption{(a) An Illustration of how to compute $\epsilon$ and $h$. Here $\mathbf{d}(\mathbf{x}_1)$ and $\mathbf{d}(\mathbf{x}_2)$ are two displacement vectors for landmark $\mathbf{x}_1$ and $\mathbf{x}_2$ respectively. $\mathit{h}=\norm{\mathbf{x}_1-\mathbf{x}_2}$ is the distance between $\mathbf{x}_1$ and $\mathbf{x}_2$. While $\epsilon=\norm{\mathbf{d}(\mathbf{x}_1)-\mathbf{d}(\mathbf{x}_2)}^2$ measures the displacement difference; (b) A hypothetical $\hat{\gamma}(h)$ generated from the vector field in Fig.1(a). Since the vector field has 5 displacement vectors, $\hat{\gamma}(h)$ has $\frac{\mathit{5}(\mathit{5}-1)}{2}=10$ value points. $\mathrm{p}$ and $\mathrm{q}$ are two value points that demonstrate the typical increasing trend of $\hat{\gamma}(h)$. Here $\mathrm{h}_\mathrm{p} < \mathrm{h}_\mathrm{q}$, thus $\epsilon_\mathrm{p}<\epsilon_\mathrm{q}$.}
	\vspace{-2mm}
\end{figure}

Fig.1(a) illustrates computing $\epsilon$ and $\mathit{h}$ for two vectors. Fig.1(b) shows a hypothetical variogram cloud generated from the vector field in Fig.1(a). The horizontal and vertical axes represent $\mathit{h}$ and $\epsilon$ respectively. Given $\mathit{K}$ displacement vectors, each $\mathbf{d}(\mathbf{x})$ has $\mathit{K}-1$ pairs of corresponding $(\mathit{h}_\mathit{ij},\epsilon_\mathit{ij})$, thus $\hat{\gamma}(h)$ contains $\frac{\mathit{K}(\mathit{K}-1)}{2}$ value points.

A common dependency assumption is that displacement vectors which are close to each other tend to be more similar than those far apart \cite{VarioBook}. In other words, for a point in $\hat{\gamma}(h)$, a smaller $\mathit{h}$ usually corresponds to a smaller $\epsilon$. As a result, a typical $\hat{\gamma}(h)$ tends to exhibit an increasing trend. For conciseness, In the rest of this article, we call $\hat{\gamma}(h)$ variogram.

		\vspace{-2mm}
\subsection{Potential FLEs}

A pair of annotated landmarks $(\mathbf{x},\mathbf{x}')$ forms a displacement vector $\mathbf{d}(\mathbf{x})$, which should indicate the deformation of $\mathbf{x}$. If $\mathbf{d}(\mathbf{x})\in\mathcal{D}$ exhibits a spatial dependency that differs from other vectors, it could indicate FLE for $(\mathbf{x},\mathbf{x}')$. We call these abnormally behaved vectors \textit{outliers}. In general, there are global outliers $\lambda_\mathrm{G}$ and local outliers $\lambda_\mathrm{L}$: 
\begin{itemize}
	\item [$\lambda_\mathrm{G}$:] have large differences with the majority of displacement vectors in $\mathcal{D}$.
	\item [$\lambda_\mathrm{L}$:] do not have extreme values, but are considerably different from their neighbors. 
\end{itemize}

Vector outliers tend to have different spatial dependence from other landmarks, which can be captured by the values of $(\mathit{h},\epsilon)$, hence we can use $\hat{\gamma}(h)$ to distinguish $\lambda_\mathrm{G}$ and $\lambda_\mathrm{L}$. In the example of Fig.2(a), we deliberately added two problematic landmarks, one with global error $\lambda_\mathrm{G}$ (blue) and one with local error $\lambda_\mathrm{L}$ (green), to a vector field. In Fig.2(b), all blue points in $\hat{\gamma}(h)$ are from adding $\lambda_\mathrm{G}$, which can be easily identified because all of its corresponding points have distinctively higher values of $\epsilon$. In Fig.2(c), all green points in $\hat{\gamma}(h)$ are from adding $\lambda_\mathrm{L}$. We can also distinguish $\lambda_\mathrm{L}$ at the bottom-left corner of $\hat{\gamma}(h)$, because some of its points yield small $\mathit{h}$ while having unusually large $\epsilon$. 

\begin{figure}[H]
	\centering
		\vspace{-5mm}
	\includegraphics[height=3.0cm]{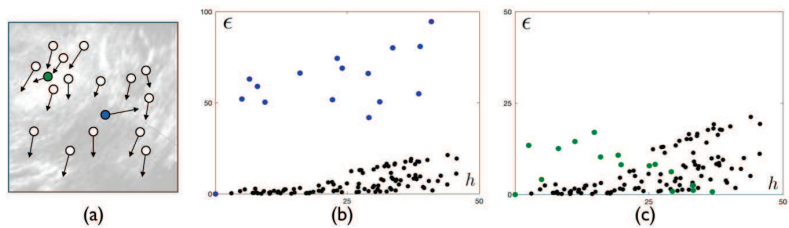}
	\vspace{-3mm}
	\caption{(a) Manually added atypically behaved displacement vectors. The blue point is $\lambda_\mathrm{G}$, the green point is $\lambda_\mathrm{L}$ (b,c) Colorized global and local outliers identified in $\hat{\gamma}(h)$. Each point represents the difference between a pair of displacement vectors, e.g., blue points are vector pairs that involve the global outlier. }
	\vspace{-5mm}
\end{figure}

It is noteworthy that some critical displacement vectors, which indicate large tissue deformation, may share the same features as outliers. Therefore, $\hat{\gamma}(h)$ is used to flag suspicious $\lambda_\mathrm{G}$ and $\lambda_\mathrm{L}$, which can be further examined by experienced radiologists.

\subsection{Atypical variogram patterns}
Since $\mathit{h}$ represents the distance between a pair of vectors, the distribution of annotated landmarks can also be reflected in the pattern of $\hat{\gamma}(h)$. Compared to observing a 3D visualization of $\mathcal{D}$, where atypical patterns may be hidden because of the viewpoint,, the variogram's 2D representation provides a clearer representation of landmark distribution. Fig.3(a) shows the typical smooth and steadily increasing pattern of an evenly distributed vector field $\mathcal{D}$. Other variogram patterns may indicate undesirable distributions of the vector field. Two undesirable patters are clustered landmarks and isolated landmarks:

\begin{enumerate}
	\item If landmarks are clustered into two (or more) distinct groups, the clustering is evident in $\hat{\gamma}(h)$ as illustrated in Fig. 3(b).
	\item If a landmark is isolated from other landmarks, its points in $\hat{\gamma}(h)$ only exist in areas where $\mathit{h}$ is large. Fig.3(c) shows an isolated landmark and its values in $\hat{\gamma}(h)$.
\end{enumerate}

\begin{figure}[t]
	\centering
	\includegraphics[width=11.5cm]{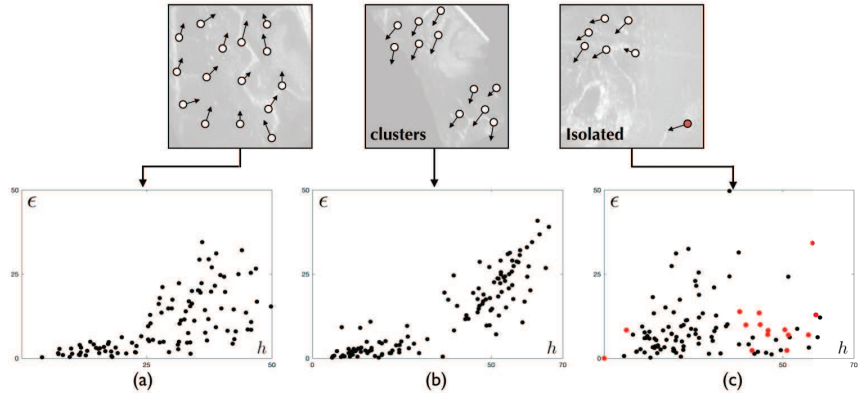}
	\vspace{-5mm}
	\caption{(a) $\hat{\gamma}(h)$ of an evenly distributed vector field; (b) $\hat{\gamma}(h)$ of a vector field that has clusters; (c) $\hat{\gamma}(h)$ of a vector field that has an isolated point. }
	\vspace{-4mm}
	\end{figure}

We construct $\hat{\gamma}(h)$ for all data and manually flag cases with the above atypical patterns for further examinations.

\section{Experiments}

RESECT \cite{RESECT} and BITE \cite{BITE} are two high-quality clinical datasets that contain \textit{p}-MR and \textit{i}-US images of patients with brain tumors. They are widely used to evaluate registration algorithms for image-guided neurosurgery. RESECT includes 23 patients with each having \textit{p}-MR (\textit{p}MR), before-resection US (bUS), during-resection US (dUS) and after-resection US (aUS) images. Four image pairs, i.e., \textit{p}MR-aUS, \textit{p}MR-bUS, bUS-aUS and bUS-dUS, have been annotated with corresponding landmarks. For BITE, pre- and post-operative MR, and \textit{i}-US images have been acquired from 14 patients. These images were further annotated and put into three groups (1) Group 1: bUS and aUS; (2) Group 2: \textit{p}MR-bUS; (3) Group 4: \textit{p}MR and post-MR.  

In order to provide an objective, thrid-party screening of the annotations in these two datasets, we generated $\hat{\gamma}(h)$ for all 700+ landmark pairs and flagged those landmark pairs with potential FLE issues. Two operators visually inspected the flagged landmark pairs and together categorized them into three categories: (1) They are certain that the landmark pair is problematic;  (2) $\hat{\gamma}(h)$ looks atypical, but they are unsure whether the landmark pair is problematic; (3) $\hat{\gamma}(h)$ looks normal. In addition, they also flagged cases with clusters or isolated landmark $\hat{\gamma}(h)$ problems.

\vspace{-2mm}

\subsubsection{Findings} After the \textit{objective screening}, we found that the vast majority of landmarks have normal-looking $\hat{\gamma}(h)$, which indicates that both datasets have high-quality annotations. In total, there are 29 pairs of landmarks that potentially have FLEs. In addition, we also identified 4 cluster cases and 11 isolated landmarks. All flagged data are summarized in Table 1.  Fig.4 and Fig.5 show the $\hat{\gamma}(h)$ of some landmark pairs that were flagged as potentially having FLE's. Fig.6 gives two examples of $\hat{\gamma}(h)$'s of a flagged \textit{cluster} and isolated landmarks. 

\begin{figure}[t]
	\vspace{-4mm}
	\centering
	\includegraphics[width=11cm]{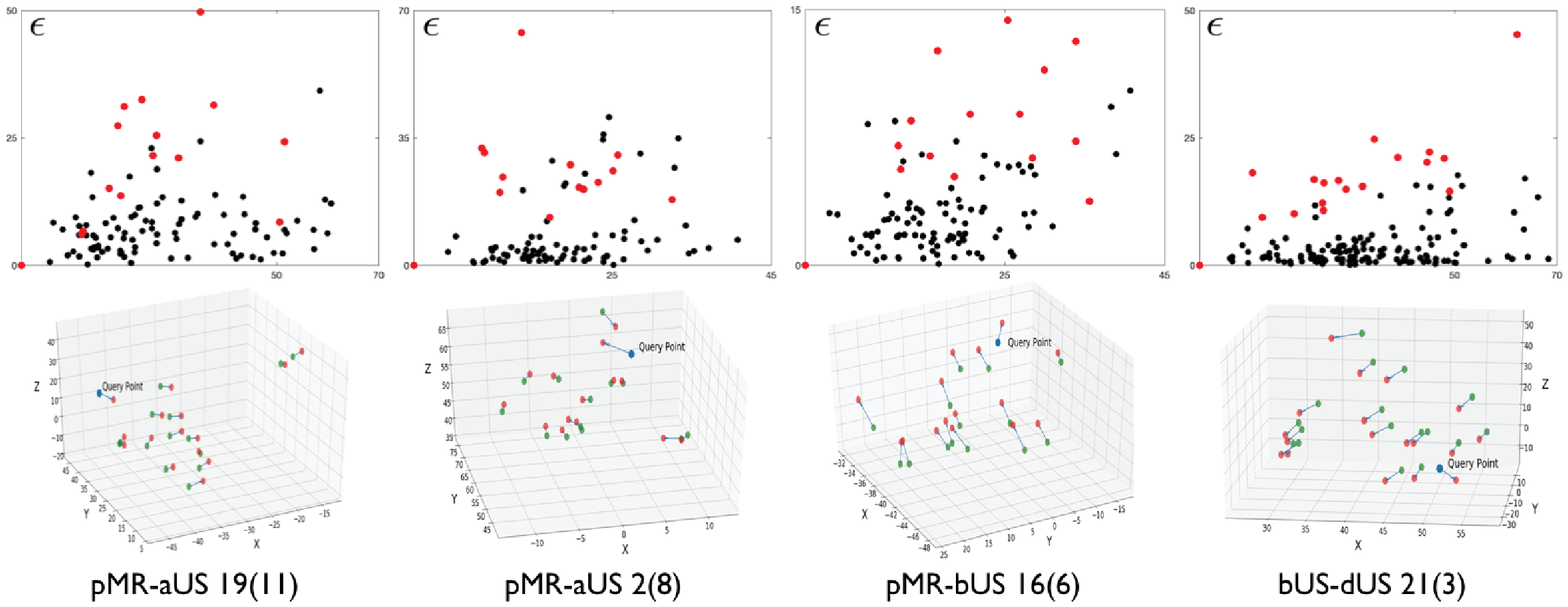}
	\vspace{-2mm}
	\caption{Examples of category one atypical landmark pairs (red). The first row shows their $\hat{\gamma}(h)$'s while the second row displays their 3D displacement vectors. }
	\label{fig:correct1}
	\vspace{-5mm}
\end{figure}

\begin{table}[H]
	\centering
	\vspace{-4mm}
	\begin{tabular}{|c|c|c|c|c|l|c|l|}
		\hline
		\multicolumn{2}{|l|}{}                            & Certain                                                              & Unsure                                                                       & \multicolumn{2}{l|}{Cluster} & \multicolumn{2}{l|}{Isolated}                                                      \\ \hline
		\multirow{4}{*}{RESECT} & \textit{p}MR-aUS                 & \begin{tabular}[c]{@{}c@{}}1(9), 2(10), 19(11) \\ 2(8)\end{tabular}  & \begin{tabular}[c]{@{}c@{}}3(5), 4(3), 7(1, 4),\\ 14(1),\end{tabular}        & \multicolumn{2}{c|}{n/a}     & \multicolumn{2}{c|}{18(14)}                                                        \\ \cline{2-8} 
		& \textit{p}MR-bUS                 & 1(9), 16(6)                                                          & \begin{tabular}[c]{@{}c@{}}1(13), 2(14), 3(1) \\ 15(13), 25(15)\end{tabular} & \multicolumn{2}{c|}{19}      & \multicolumn{2}{c|}{2(14), 19(1)}                                                  \\ \cline{2-8} 
		& bUS-aUS                 & \begin{tabular}[c]{@{}c@{}}1(7), 7(8), 12(11) \\ 24(14)\end{tabular} & 15(3)                                                                        & \multicolumn{2}{c|}{25}      & \multicolumn{2}{c|}{\begin{tabular}[c]{@{}c@{}}1(11), 18(12)\\ 19(1)\end{tabular}} \\ \cline{2-8} 
		& bUS-dUS                 & 21(3), 27(11)                                                 & 6(10), 7(22)                                                                 & \multicolumn{2}{c|}{19}      & \multicolumn{2}{c|}{n/a}                                                           \\ \hline
		\multirow{3}{*}{BITE}   & \multicolumn{1}{l|}{G1} & 3(4), 10(1)                                                          & n/a                                                                          & \multicolumn{2}{c|}{12}      & \multicolumn{2}{c|}{\begin{tabular}[c]{@{}c@{}}2(3), 4(4)\\ 9(6)\end{tabular}}     \\ \cline{2-8} 
		& \multicolumn{1}{l|}{G2} & 9(5)                                                                 & 12(1)                                                                        & \multicolumn{2}{c|}{n/a}     & \multicolumn{2}{c|}{3(21)}                                                         \\ \cline{2-8} 
		& \multicolumn{1}{l|}{G4} & n/a                                                                  & 1(6)                                                                         & \multicolumn{2}{c|}{n/a}     & \multicolumn{2}{c|}{3(16)}                                                         \\ \hline
	\end{tabular}
	\vspace{+3mm}
	\caption{Indexes of problematic landmark pairs, e.g., 1(9) means patient 1 and landmark pair No.9.}
	\vspace{-15mm}
\end{table}

Since the brain may undergo deformation during surgery, atypical behaviors of $\mathbf{d}$'s may indicate actual deformation of the brain. In order to investigate whether ``problematic" landmarks contain localization errors, we sent the findings (mixed with good landmarks) to 3 experienced neuro-radiologists for validation and rating. They carefully examined the landmark coordinates in physical space using Slicer \cite{Slicer} and assigned a score from [1(poor), 2(questionable), 3 (acceptable) , 4 (good)].  Landmarks in Category one and category two received an average score of 1.5 and 2.4 respectively. Fig.7(a) shows the user interface for the validation procedure.

\begin{figure}[t]
	\centering
	\vspace{-3mm}
	\includegraphics[width=11cm]{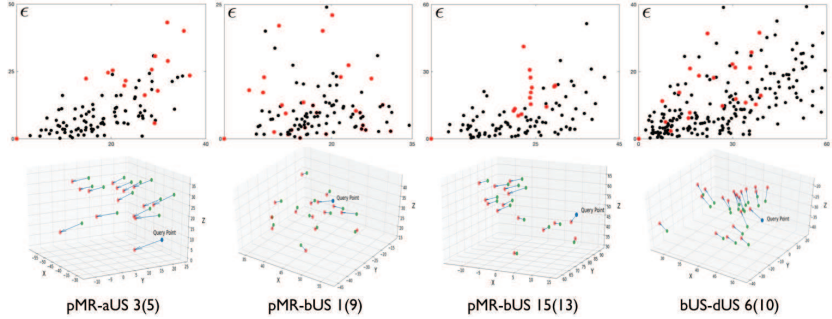}
	\vspace{-2mm}
	\caption{Examples of category two landmark pairs. These landmarks have atypical $\hat{\gamma}(h)$'s, but their displacement vectors could be reasonable. }
	\vspace{+3mm}
\end{figure}

\begin{figure}[t]
	\centering
	\includegraphics[width=11cm]{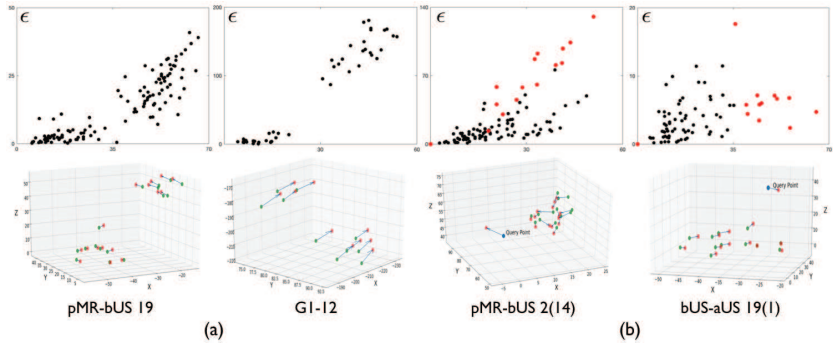}
	\vspace{-3mm}
	\caption{Examples of (a) two flagged \textit{clusters} and (b) two isolated landmarks. }
	\vspace{-4mm}
\end{figure}

\vspace{-2mm}
\subsubsection{Potential evaluation bias} FLEs and non-evenly distributed landmarks can incur bias in the registration evaluation:

\begin{enumerate}
	\item Since most FRE metrics takes into account all landmarks equally (not weighted), landmarks with FLEs produce false registration error and can drive the algorithm towards aligning those inaccurately located markers.
	\item \textit{cluster} and isolated landmarks both are not (densely) evenly distributed, thus they incur evaluation bias that prioritizes regions that have landmarks. As in the \textit{p}-US to \textit{i}-US registration example shown in Fig.7(b), landmarks only exist in the sulcus region (as two clusters). Here, we have registered \textit{p}-US images $I_\mathrm{r}$ and $I_\mathrm{r}^*$, which are registered by two different registration methods. In the landmark-based evaluation, $I_\mathrm{r}$ has a better FRE score than $I_\mathrm{r}^*$ because it perfectly aligns the sulcus region (while ignoring the rest of image). However, in surgeons' eyes, it is $I_\mathrm{r}^*$ which is more reasonable (useful) since it provides accurate tumor boundary alignment. 
\end{enumerate}


\begin{figure}[t]
	\centering
	\includegraphics[width=12.2cm]{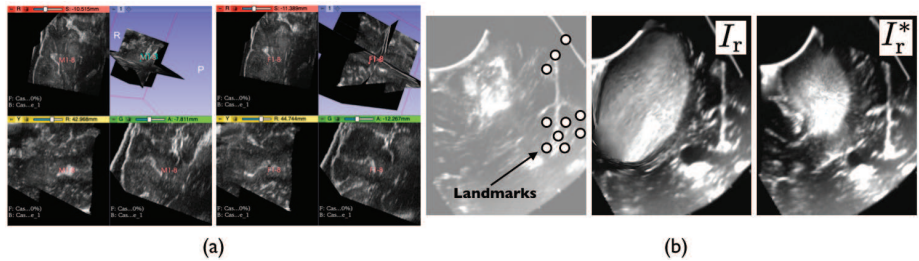}
	\vspace{-6mm}
	\caption{ (a) The user interface for the landmark validation; (b) Evaluation bias caused by un-evenly distributed landmarks. On the left is an \textit{i}-US image. $I_\mathrm{r}$ and $I_\mathrm{r}^*$ are registered \textit{p}-US's using two different registration methods. }
	\vspace{-5mm}
\end{figure}

	\vspace{-4mm}
\section{Discussion}
	\vspace{-2mm}
Manual landmark annotation is (mostly) a subjective task, thus public datasets may inevitably have FLEs. To mitigate the evaluation bias caused by FLEs, one strategy is to apply a landmark weighting (selection) scheme \cite{Fitzpatrick5,Shamir,Thompson}. However, at the current stage, these methods are still not thoroughly validated and should be approached with caution. 
RESECT and BITE are both purposefully created for image-guided neurosurgery. We understood that anatomical features in the brain, mainly corner points and small holes, will most likely not appear uniformly in image space. In order to achieve a non-biased comparison between registration algorithms, dataset providers can add notes to describe the distribution (limitations) of landmarks so that users, if necessary, can incorporate more criteria, such as surfaces \cite{Santos}, in the evaluation.

Whether or not public datasets assure unbiased comparisons for registration evaluation is a crucial question that deserves more attention from the image registration community. From this \textit{objective third-party screening}, we conclude that RESECT and BITE are both reliable datasets with a small number of problematic landmarks and landmark distributions that may bias the non-weighted FRE evaluation. Besides the fore-mentioned using advanced evaluation criteria, some data providers offer to update their repository based on users' feedback, e.g., adding or correcting annotations, which is another solution to assure unbiased registration evaluation. 

As a tool for \textit{objective screening}, variogram may have limitations. Nevertheless, we believe that this paper will serve as a foundation and draw more attenton to this topic.

%
%

\end{document}